\documentclass{article}

     \usepackage[preprint]{neurips_2022}

\usepackage{xr-hyper}

\usepackage[utf8]{inputenc} % allow utf-8 input
\usepackage[T1]{fontenc}    % use 8-bit T1 fonts
\usepackage{hyperref}       % hyperlinks
\usepackage{url}            % simple URL typesetting
\usepackage{booktabs}       % professional-quality tables
\usepackage{amsfonts}       % blackboard math symbols
\usepackage{nicefrac}       % compact symbols for 1/2, etc.
\usepackage{microtype}      % microtypography
\usepackage{xcolor}         % colors

\usepackage[american]{babel}
\usepackage{mathtools} % amsmath with fixes and additions
\usepackage{siunitx} % for proper typesetting of numbers and units
\usepackage{tikz} % nice language for creating drawings and diagrams

\usepackage{graphicx}
\usepackage{amsfonts,bm}
\usepackage{subcaption}
\usepackage{floatrow}
\usepackage{authblk}
\hypersetup{colorlinks=true,linkcolor=blue!80,urlcolor=purple!80,citecolor=blue!80}
\newfloatcommand{capbtabbox}{table}[][\FBwidth]

\makeatletter
\newcommand*{\addFileDependency}[1]{% argument=file name and extension
  \typeout{(#1)}
  \@addtofilelist{#1}
  \IfFileExists{#1}{}{\typeout{No file #1.}}
}
\makeatother

\newcommand*{\myexternaldocument}[1]{%
    \externaldocument{#1}%
    \addFileDependency{#1.tex}%
    \addFileDependency{#1.aux}%
}

\myexternaldocument{supplement}

\title{Machines Explaining Linear Programs}

\author{\textbf{David Steinmann}\textsuperscript{\rm 1,$\dagger$} \quad
\textbf{Matej Zečević}\textsuperscript{\rm 1}
\quad \textbf{Devendra Singh Dhami}\textsuperscript{\rm 1,3} \quad \textbf{Kristian Kersting}\textsuperscript{\rm 1,2,3}}
\affil{\textsuperscript{\rm 1}Computer Science Department, TU Darmstadt, \textsuperscript{\rm 2}Centre for Cognitive Science, TU Darmstadt,\\
\textsuperscript{\rm 3}Hessian Center for AI (hessian.AI), \textsuperscript{\rm $\dagger$}correspondence:\ \texttt{david.steinmann@tu-darmstadt.de}\\
\vspace{-.5cm}
}
\author{%
  David S.~Hippocampus\thanks{Use footnote for providing further information
    about author (webpage, alternative address)---\emph{not} for acknowledging
    funding agencies.} \\
  Department of Computer Science\\
  Cranberry-Lemon University\\
  Pittsburgh, PA 15213 \\
  \texttt{hippo@cs.cranberry-lemon.edu} \\
}

\DeclareCaptionLabelFormat{andtable}{#1~#2  \&  \tablename~\thetable}

\begin{document}

\maketitle

\begin{abstract}
  There has been a recent push in making machine learning models more interpretable so that their performance can be trusted. Although successful, these methods have mostly focused on the deep learning methods while the fundamental optimization methods in machine learning such as linear programs (LP) have been left out. Even if LPs can be considered as whitebox or clearbox models, they are not easy to understand in terms of relationships between inputs and outputs. As a linear program only provides the optimal solution to an optimization problem, further explanations are often helpful. In this work, we extend the attribution methods for explaining neural networks to linear programs. These methods explain the model by providing relevance scores for the model inputs, to show the influence of each input on the output.
  Alongside using classical gradient-based attribution methods we also propose a way to adapt perturbation-based attribution methods to LPs. Our evaluations of several different linear and integer problems showed that attribution methods can generate useful explanations for linear programs. However, we also demonstrate that using a neural attribution method directly might come with some drawbacks, as the properties of these methods on neural networks do not necessarily transfer to linear programs. The methods can also struggle if a linear program has more than one optimal solution, as a solver just returns one possible solution. Our results can hopefully be used as a good starting point for further research in this direction.
\end{abstract}

\section{Introduction}
Linear Programs (LP) are a commonly used technique for optimization, widely applied in schedule optimization in smart grids \citep{zhu2012integer}, to solve combinatorial problems \citep{paulus2021comboptnet} and most recently resource planning during Covid-19 crisis \citep{petrovic2020simulation}. Besides their applicability to a wide range of problems, their efficiency is an important reason for their popularity \citep{spielman2004smoothed}. Although very popular, there have not been many efforts to make their results understandable, especially for non-experts in optimization. The only major effort in this direction is sensitivity analysis \citep{higle2003sensitivity}, which is commonly used in conjunction with LPs and provides information about how much the solution changes if the input parameters are modified slightly \citep{ward1990approaches}. 
% Even though this information is helpful, it is not really sufficient to fully understand a complex LP.

In recent years, explainable artificial intelligence (XAI) methods have risen in popularity \citep{adadi2018peeking}. Providing explanations for a machine learning (ML) model is useful in many different ways. For experts, an explanation allows validating if the behavior of the model is correct, or if it just follows a Clever Hans behavior \citep{stammer2021right}. Explaining the model's reasoning also helps laymen to build trust in the model as it is easier to comprehend and trust the model's decision \citep{gunning2019xai}. XAI may also facilitate the knowledge of humans who work with the ML system. One category of XAI is attribution methods. These provide contribution scores for all input features of a given ML model thus highlighting important and irrelevant features. 
There are various attribution methods such as integrated gradients \citep{sundararahan2017axiomatic}, Grad-CAM \citep{selvaraju2017gradcam} or LIME \citep{riberio2016why}. They are mostly focused on deep neural networks and especially image-based tasks, as it is possible to visualize the attributions as an overlay over the original image, highlighting important segments. As common as XAI, and specifically attribution methods, are for deep neural networks, this topic has not been discussed for LPs. As the integration of optimizers and deep learning has gained significant interest \citep{amos2017optnet,paulus2021comboptnet} it is natural to apply these methods to other types of models.

\begin{figure}
    \centering
    \includegraphics[width=\textwidth]{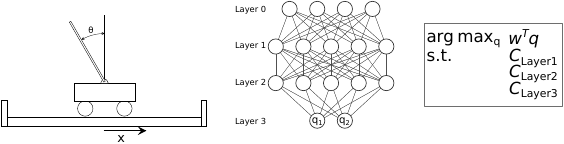}
    \caption{The inverted pendulum on a cart problem \textbf{(left)} can be solved as a ReLU network \textbf{(middle)} and can be explained via attribution methods. It is also possible to encode the ReLU network as a linear program via defining constraint for each layer \textbf{(right)}. However, even if the LP formulation is equivalent to the ReLU-net, no equivalent explanation methods exist.}
    \label{fig:relu_net}
\end{figure}

In this work, we extend the attribution methods to obtain \emph{explanations for LPs}, which we term XLP, that share equivalences with neural nets (NN) and further justify the transfer. It has been shown that ReLU NN \citep{nair2010rectified} or even recurrent NN \citep{rumelhart1986learning} might be viewed as LPs \citep{wang1992recurrent} (see Fig. \ref{fig:relu_net}). We note that several attribution methods that require a certain model architecture cannot be applied to linear programs directly as the structure of an LP is different from the structure of a neural network. This makes all propagation-based approaches not feasible for LPs. Additionally, attribution methods that require an error measurement of the model cannot be applied directly, as an LP does not have an inherent model error. We make use of four different attribution methods: saliency maps, gradient times input, integrated gradients, and a modified version of occlusion to explain LPs, by computing attributions either for the optimal solution of the linear program or for some function applied to the optimal solution. As an important result, we show that standard properties of attribution methods, namely sensitivity, completeness, and implementation invariance do not hold for LPs. \label{limitations1}

Overall, we make the following key contributions: (1) We show that neural attribution methods can be extended to explain LPs. (2) We provide an extensive analysis of the theoretical properties of these methods on LPs and show that these properties do not necessarily translate from NNs to LPs. (3) We provide an empirical evaluation of the four attribution methods on four different LP use cases. (4) Furthermore, we provide an empirical evaluation of a real-world optimization problem and a MAP inference problem.

\section{Background and Related Work}

{\bf Linear Programs.} LPs are a special family of optimization programs, restricted to a linear objective function and linear constraints \citep{karloff2008linear}. An LP can be formulated as
\begin{equation} \label{eq:lp}
\begin{aligned}
    \mathbf{x}^* =& \arg\max\nolimits_{\mathbf{x}} & \mathbf{w}^T \mathbf{x} \\
    & \mathbf{s.t.} & \mathbf{A} \mathbf{x} \leq \mathbf{b}
\end{aligned}
\end{equation}
where $\mathbf{x} \in \mathbb{R}^n$ is the optimization variable and $\mathbf{w} \in \mathbb{R}^n$ is weight vector. The optimization constraints are specified by the matrix $\mathbf{A} \in \mathbb{R}^{m \times n}$ and the vector $\mathbf{b} \in \mathbb{R}^m$. $\mathbf{A}$, $\mathbf{b}$ and $\mathbf{w}$ are constant parameters of the LP. $\mathbf{x}^*$ is called the optimal solution of the LP. Integer programs (ILP) add an additional restriction requiring $\mathbf{x}$ to be a valid \emph{integer} points.
For an LP, the solution map $\mathbf{S}$ is a function that maps from the parameters to the optimal solution. It is defined as:
$\mathbf{S}: \mathbf{A}, \mathbf{b}, \mathbf{w} \rightarrow \mathbf{x}^*$. $\mathbf{S}$ can be differentiated thereby allowing for the computation of gradients of an LP or ILP \citep{paulus2021comboptnet, agrawal2019differentiable}. Similarly, it is possible to define the objective map $\mathbf{O}$ as the function which maps from the parameters to the objective function value of the optimal solution.

% \begin{equation}
% \begin{alignedat}{3}
%     \mathbf{O}: \quad &\mathbf{A}, \mathbf{b}, \mathbf{w} \ &&\rightarrow \ &&\mathbf{w}^T\mathbf{x}^*\\
%     \mathbf{O}(&\mathbf{A}, \mathbf{b}, \mathbf{w}) \ &&= \ &&\mathbf{w}^T \mathbf{S}(\mathbf{A}, \mathbf{b}, \mathbf{w})
% \end{alignedat}
% \end{equation}

{\bf Sensitivity Analysis.}  Sensitivity analysis is a general method to explain the behavior of a mathematical model to variations in its input parameters \citep{saltelli2010avoid}. A problem is \emph{sensitive} to an input parameter if changing this parameter slightly also changes the solution of the model. In terms of LPs, it is investigated how much the parameter values of $\mathbf{w}$ or $\mathbf{b}$ can be changed such that the current optimal basis remains optimal \citep{koltai2000difference}. It is also reported how fast the optimal solution value changes within these intervals. The results often depend on the used LP solver, making them difficult to interpret correctly \citep{jansen1997sensitivity}. Sensitivity analysis differs from the approach of this work in the sense that we aim to find the influence of inputs on the LP output, not the input sensitivity.

%Sensitivity analysis is mostly performed for the objective function weights $\mathbf{w}$ and the right-hand side coefficients $\mathbf{b}$. The results of sensitivity analysis often depend on the used LP solver, which can make it difficult to interpret them correctly \citep{jansen1997sensitivity}. In general, sensitivity analysis provides different insights than our approach in this work. Here, an input value is considered important if it has a high influence on the current output of the LP, but not based on the sensitivity of the LP on this parameter.

{\bf Neural attribution methods.} Given a model $\mathbf{M}$ with input $\mathbf{x} = (\mathbf{x}_1, \cdots, \mathbf{x}_n)$, an attribution method provides attribution scores $\mathbf{c}_\mathbf{M}(\mathbf{x}) = (\mathbf{c}_1, \cdots, \mathbf{c}_n)$, such that each $\mathbf{c}_i$ describes the relevance of $\mathbf{x}_i$ on the output of $\mathbf{M}$. %These can for example be categorized into gradient based, perturbation based and causality based methods \citep{ancona2018towards}.

Saliency maps \citep{simonyan2014deep} are one of the simplest and earliest examples of gradient-based methods where the backpropagated gradients at the input values are used as attributions. Guided backpropagation \citep{springenberg2015striving} and deconvolutional nets \citep{zeiler2014visualizing} both have a slightly different approach by masking out some of the gradient signals. Instead of propagation the gradients back through the whole network to the input layer, Grad-CAM uses the gradient signals at the last convolutional layer to combine both high-level information and provide spatially accurate attributions \citep{selvaraju2017gradcam}. These methods can be considered \emph{local}, as they are based solely on the gradients. On the other hand, \emph{global} methods describe the effect of a feature on the output w.r.t.\ a baseline, by multiplying the gradient with the model input. Examples are gradient times input \citep{shrikumar2016not} or integrated gradients \citep{sundararahan2017axiomatic}.

%As all the previous methods generate attributions just based on the gradients, they can be considered \emph{local} methods. Methods that generate attributions by computing a gradient-based quantity and multiplying with the model input can be considered \emph{global} methods, as they describe the effect of a feature on the output w.r.t.\ a baseline. Examples for these methods are gradient times input \citep{shrikumar2016not} or integrated gradients \citep{sundararahan2017axiomatic}. 

Layerwise relevance propagation \citep{bach2015pixel}, deep Taylor decomposition \citep{montavon2017explaining} and DeepLIFT \citep{shrikumar2017learning} are all examples of \emph{propagation-based} attribution methods. They compute the attributions for the input values by propagating relevance backwards through the network and explicit rules allow for better attribution control \citep{shrikumar2017learning}.
\emph{Perturbation-based} approaches alter, mask or remove some input features of the model. Attributions are then obtained by comparing the original model output with the output of the changed features e.g., Occlusion \citep{zeiler2014visualizing} or LIME \citep{riberio2016why}. They can be applied to any model without knowledge about the model architecture but tend to get slower as the input size grows \citep{ivanovs2021perturbation}.

Generating attributions can also be approached from a \emph{causal} point of view. It is possible to use the notion of causality \citep{pearl2009causality} to interpret a neural network as a structural causal model \citep{chattopadhyay2019neural}. This allows computing the average causal effects of input neurons on the output neurons. This inherently causal metric can be used as attribution. CXPlain \citep{schwab2019cxplain} provides a framework to generate attributions for an arbitrary model based on the principle of \emph{Granger} causality \citep{granger1969investigating}. Similar to a perturbation-based method, they compute the attributions as a difference between the original model output and the model with a masked input feature. % To avoid the problem of computational inefficiency as the input size grows, \citet{schwab2019cxplain} propose to train a model which can predict the attributions.

\section{XLP Using Attribution Methods}
Although attribution methods have their flaws \citep{kindermans2019reliability}, they have proven to be effective, especially for image-based tasks on neural networks. The concept of attributions can be directly translated to LPs, specifically to the solution and the objective map functions. For the solution map, the attributions show the influence of all parameters on the optimal solution. When computed for the objective map, they show the influence of the parameters on the scalar output of the objective function.

This view of an LP makes it possible to apply attribution methods in general. However, not all neural attribution methods are suitable for LPs such as those requiring an NN-like structure of the model (e.g. convolutional layers as for Grad-CAM), as such structures are not available in an LP. Additionally, propagation-based approaches can also not be applied directly. The solution map of an LP is a single, non-linear function and it is not possible to propagate attributions step-by-step through this function as opposed to a NN. We identify and present four different LP-applicable methods.

{\bf Saliency:} Using saliency maps is one of the simplest approaches to generate attributions via gradients. The attributions for all input parameters are the partial derivatives of that parameter w.r.t. the model function \citep{simonyan2014deep}. For an LP $\mathbf{M}$, the saliency attributions are:
\begin{equation} \label{saliency}
\mathbf{c}_\mathbf{M}(\mathbf{x}_i) = \frac{\partial \mathbf{M}(\mathbf{x})}{\partial \mathbf{x}_i}
\end{equation}
where $\mathbf{x}_i$ is one of $\{\mathbf{A}, \mathbf{b}, \mathbf{w}\}$ and the model can be either $\mathbf{S}(\mathbf{A}, \mathbf{b}, \mathbf{w})$ or $\mathbf{O}(\mathbf{A}, \mathbf{b}, \mathbf{w})$. 

{\bf Gradient times Input (GxI):} There have been arguments that multiplying the partial derivative of a parameter with the parameter itself increases the quality of the attribution \citep{shrikumar2016not}. GxI is an extension of the saliency method in this direction. The partial derivatives of the parameters w.r.t.\ model function are multiplied with the parameter values themselves, resulting in the following formulation, with the same elements as Eq. \ref{saliency}:
\begin{equation}
\mathbf{c}_\mathbf{M}(\mathbf{x}_i) = \mathbf{x}_i \frac{\partial \mathbf{M}(\mathbf{x})}{\partial \mathbf{x}_i}
\end{equation}

{\bf Integrated Gradients (IG):} To obtain the IG attributions, it is necessary to select a \emph{baseline} point in the input space of the model which acts as a reference to measure the relevance against. In the context of an LP, the baseline $\bar{\mathbf{x}}$ consists of the values $\bar{\mathbf{A}}$, $\bar{\mathbf{b}}$ and $\bar{\mathbf{w}}$. The attributions are obtained by integrating the model gradient along the path between baseline and input value.
\begin{equation}
\mathbf{c}_\mathbf{M}(\mathbf{x}_i) = (\mathbf{x}_i - \bar{\mathbf{x}}_i) \cdot \int_{\alpha = 0}^{1} \frac{\partial S(\tilde{\mathbf{x}})}{\partial \tilde{\mathbf{x}}_i} \Biggr|_{\tilde{\mathbf{x}} = \bar{\mathbf{x}} + \alpha(\mathbf{x} - \bar{\mathbf{x}})} d\alpha 
\end{equation}
This method was specifically designed so that the attributions fulfill several desirable properties, which encouraged the evaluation of the theoretical capabilities of attribution methods. The behavior of attribution methods on LPs w.r.t.\ these properties is analyzed in more detail in section \ref{sec:properties}. When this method is applied to classification tasks with NNs, a reference point should be \emph{neutral} in terms of the classification output \citep{bach2015pixel}. This approach is not directly suitable for LPs, as they have no classification output. Therefore, we use baseline points corresponding to a neutral problem state, which are selected based on the general knowledge about the problem.
% \begin{figure}
%     \centering
%     \includegraphics{lp_masking.pdf}
%     \caption{This figure shows the effects of masking a single element of one parameter from the problem. Setting a single entry of $\mathbf{b}$ to zero also affects the corresponding values of $\mathbf{A}$ (marked in red). If the full constraint is removed at once, no undesired side-effects occur and it is possible to compute attributions for the constraint. }
%     \label{fig:lp_masking}
% \end{figure}

{\bf Occlusion (Occ):} This is a simple perturbation-based attribution method \citep{zeiler2014visualizing}% directly connected to the idea of \emph{Granger causality}\footnote{For an extended discussion, see Appendix \ref{sec:gc_lp}.}. 
 and computes the attributions for a feature $\mathbf{x}_i$ under the model $\mathbf{M}$ as:
$
\mathbf{c}_\mathbf{M}(\mathbf{x}_i) = \mathbf{M}(\mathbf{x}) - \mathbf{M}(\mathbf{x}_{[\mathbf{x}_i = 0]}).
$
Intuitively, attributions are defined as the difference between the normal model output and the model output if the feature is set to zero. This assumes that an input feature with a value of zero does not influence the model outcome, which allows us to estimate the true influence of the feature. However, setting a parameter to zero in an LP does not always result in a zero \emph{influence} of that parameter. If part of $\mathbf{b}$ or $\mathbf{w}$, the change can modify the problem significantly (Fig. \ref{fig:lp_masking} in supplementary). 

We approach the problem from a slightly different perspective. Instead of masking out a single value of a parameter, which has a possibly large influence on other parameters, certain structures are masked out. These structures should be meaningful units by themselves and can be part of multiple parameters. These structures can be selected so that they do not have an undesired influence on other parts of the problem, making it possible to compute their influence.

This approach has two significant advantages: (1) The meaningful structures can be selected flexibly, whereas the other methods always provide attributions for the parameters. (2) It can be easier to understand the attributions for meaningful structures, as these already have meaning themselves. The standard parameters of an LP also have a meaning, but it might be more difficult to understand if one is not familiar with LPs. However, these advantages also come with a drawback. To compute attributions, it is necessary to specify the meaningful structures beforehand, the other methods can be directly applied to an LP in a more ``exploratory" way. Note that this approach is not restricted to occlusion, but can also be applied for other perturbation-based methods.

{\bf Computation Complexity:} In addition to the accuracy the actual runtime of an attribution method is an important factor to consider. In general, gradient-based attribution methods are known to be fast, as they require only a single forward and backward pass through the model (e.g. in case of saliency or GxI) or a fixed number of these cases (e.g. 100 to approximate the integral for IG). On the other hand, the complexity of Occ is linear in the number of selected meaningful structures, which can limit its scalability to large problems. However, computing the gradient for a large LP is time-consuming as well, so Occ with even a moderate number of structures can be faster than e.g. IG.

% We note that the selected list of candidate explanatory methods for LPs is \emph{not exhaustive}.

\section{Empirical Analysis}
\begin{figure*}[t]
    \centering
    \includegraphics[scale=0.75]{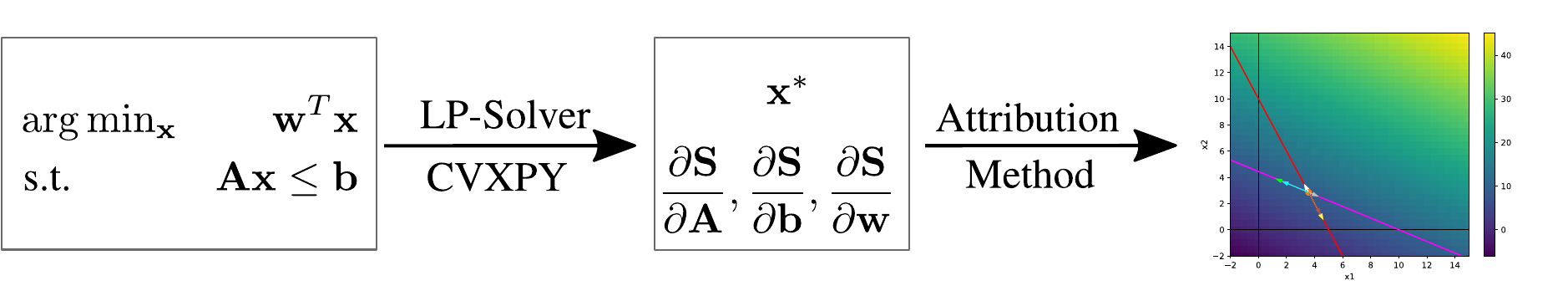}
    \caption{Process to generate attributions for a problem. We start with an LP. This LP is solved and the partial derivatives w.r.t.\ the parameters are computed with CVXPY \citep{diamond2016cvxpy}. Then, an attribution method is used to generate gradient-based attributions.}
    \label{fig:attribution_process}
\end{figure*}

We now evaluate the attribution methods on two LPs (\emph{maximum flow} (MF) and \emph{resource optimization} (RO)) and two ILPs (\emph{knapsack} (KS) and \emph{shortest path} (SP)), each with several cases. These problems were selected to cover the typical use case of linear programs, resource optimization, as well as some typical combinatorial problems. This evaluation should check if the attribution methods can generate reasonable attributions for LPs and under which circumstances they might fail (Fig. \ref{fig:attribution_process}).

\subsection{Evaluation Structure: Type Categorization}
Evaluating attribution methods is \emph{not trivial}, as there are no ground-truth attributions available to compare. Additionally, it is possible to argue that different sets of attributions provide reasonable explanations of a problem making their evaluation notoriously difficult. This problem has been tackled by defining metrics to measure the quality of attributions \citep{ancona2018towards}. %These metrics are often based on desirable properties of attribution methods.
Unfortunately, these metrics cannot be applied to LPs, as they either depend on removing specific elements of the problem or are designed for image-based tasks \citep{goh2020understanding}. Therefore, we follow the more informal approach to compare the produced attributions and decide how reasonable they are as often done to compare heatmaps of attributions for image-based NN tasks \citep{shrikumar2017learning}.

The concrete settings in all problems are kept small so the attributions can be evaluated by hand. The problem cases are split into three different types: \textbf{Type I:} These cases describe a typical situation of an LP, where a unique optimal solution exists. \textbf{Type II:} These cases have more than one optimal solution. \textbf{Type III:} Especially intricate cases, e.g. implicit constraints like $\mathbf{x} \geq 0$ are deciding factors.

\subsection{Exact Problem Descriptions}
The first problem is one of the earliest use-cases for linear programming, resource optimization \citep{dantzig1951maximization}. Given the resource costs and the sale prices of several different products, as well as a limit for each resource, the objective is to maximize the total revenue. The instance of this problem for our evaluation has two different products and resources. The problem is evaluated in all three case types with type III being the edge-case in which implicit constraint $\mathbf{x} \geq 0$ is relevant. The LP formulation for the problem with example values is:
\begin{equation} \label{eq:ro}
\begin{aligned}
    \mathbf{x}^* =& \arg\max\nolimits_{\mathbf{x}} & \left(\begin{array}{cc}
        1 & 2
    \end{array}\right)& \mathbf{x}& \\
    & \mathbf{s.t.} & \left(\begin{array}{cc}
        1 & 1 \\
        2 & 1
    \end{array}\right)& \mathbf{x} \leq \left(\begin{array}{c}
        10 \\
        10 
    \end{array}\right)
\end{aligned}
\end{equation}

\begin{figure}
\centering
\begin{minipage}{0.35\textwidth}
    \centering
    \includegraphics[scale=0.6]{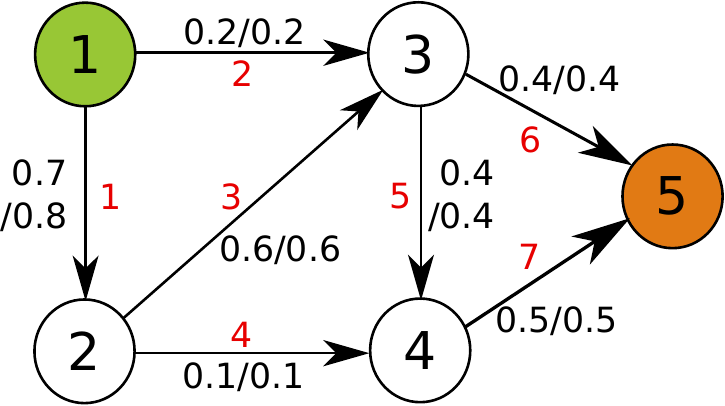}
\end{minipage}
\begin{minipage}{0.6\textwidth}
    \centering
    \captionsetup{type=table} %% tell latex to change to table
    \begin{tabular}{lll}
        \toprule
          & Description & Cases \\
        \midrule
        Type I & Unique optimal solution & RO1, RO2, RO3\\
        & & MF1, KS1, SP1 \\
        Type II & Multiple optimal solutions & RO4, MF2, KS2\\
        & & SP2 \\
        Type III & Edge cases & RO5, KS3 \\
        \bottomrule
    \end{tabular}
\end{minipage}
    \captionlistentry[table]{Evaluation cases}\label{tab:cases}
    \captionsetup{labelformat=andtable}
    \caption{\textbf{Left:} Graph layout for the maximum flow problem. The source and target nodes are highlighted in green and orange respectively. All edges have the flow/capacity for MF1 annotated. The numbers are unique edge identifiers. \textbf{Right:} Evaluation cases of the used LPs and ILPs.}
    \label{fig:mf_ex}
    
\end{figure}

The maximum flow problem is the second LP for the evaluation \citep{goldberg1988new}. Given a directed graph with designated source and target nodes as well as a maximum capacity for each edge, the objective is to maximize the flow which reaches the target. Flow can be routed along edges in the graph under two constraints: The flow along an edge cannot exceed that edge's capacity and the incoming and outgoing flow has to be the same for all nodes except source and target. The first example of this problem is type I, whereas the second is type II, as it consists of a bottleneck at the target node which results in multiple solutions of the same quality. Both cases are on the same general graph structure (Fig. \ref{fig:mf_ex}), with different edge weights between them.
%\begin{figure}
%    \centering
%    \includegraphics[scale=0.75]{mf_c1.pdf}
%    \caption{Graph layout for the maximum flow problem. The source and target nodes are highlighted in green and orange respectively. All edges have the flow and capacity for MF1 annotated. The numbers are unique identifiers for the edges.}
%    \label{fig:mf_ex}
%\end{figure}
The full mathematical formulation of the maximum flow problem can be found in Appendix \ref{sec:mf_descr}.
% and follows the LP structure of Eq. \ref{eq:lp}, where the parameter $\mathbf{A}$ encodes the graph structure.

The third problem, an ILP, is the knapsack problem \citep{salkin1975knapsack}. For a set of items with a weight and a value each, a subset with the highest value should be found which is still lighter than a maximum weight. This problem is an integer problem, as each item can either be part of the subset or not, which results in possible values of 0 and 1 for the optimization variable. The three cases of the problem cover all three types. %wherein the Type III case some items with a significantly larger weight are added to the problem. In theory, this case should not be much different from the first case, as the large and unusable item cannot be used. 
The ILP formulation of the problem follows Eq. \ref{eq:lp}. The parameter $\mathbf{A}$ is a vector of the item weights, $\mathbf{b}$ is the total capacity of the knapsack, and $\mathbf{w}$ is a vector of all item values. The knapsack problem has just a single constraint.

The last problem is the shortest path problem \citep{taccari2016integer}. Given a directed graph with a source and a target node, the objective is to find the shortest path from source to target. Each edge in the graph has an associated cost. As for MF, there are two cases with a unique optimal solution in the first and two different optimal solutions in the latter case. The graph layout is similar to Fig. \ref{fig:mf_ex}. The LP formulation and the exact graph layout can be found in Appendix \ref{sec:sp_descr}. Tab. \ref{tab:cases} summarizes the settings.

%\begin{table}
%\centering
%\begin{tabular}{lll}
%\toprule
%  & Description & Cases \\
%\midrule
%Type I & Unique optimal solution & RO1, RO2, RO3\\
%& & MF1, KS1, SP1 \\
%Type II & Multiple optimal solutions & RO4, MF2, KS2\\
%& & SP2 \\
%Type III & Edge cases & RO5, KS3 \\
%\bottomrule
%\end{tabular}
%\caption{Evaluation cases of the used LPs and ILPs.}
%\label{tab:cases}
%\end{table}

{\bf Selecting Baselines for IG.} Selecting a suitable baseline for integrated gradients on LPs is different from NNs. To select a baseline, it is recommended to use a point in the input space where the classification model has a neutral outcome. However, LPs have no classification setting. Therefore, we selected points in the input space which can be considered neutral based on general problem knowledge. An example is the baseline with equal value and weight for all items in KS, resulting in no preference for any item at the baseline. More details can be found in Appendix \ref{sec:baselines_ig}.

{\bf Meaningful Structures for Occ.} The method Occ requires the selection of problem parts for which the attributions should be computed. For the graph-based problems, all edges are considered as such. For RO, each resource, and for KS each item is used as a meaningful structure.

\subsection{Systematic Discussion of Results}
For the type I cases, the generated attributions are overall reasonable, with some notable exceptions. The attributions provided by the saliency method are skewed if the input parameters have largely different sizes such as in RO1. The parameter inputs and the results are shown in Tab. \ref{tab:res_t1}. The input has $\mathbf{b} \gg \mathbf{A}$, which results in attributions for $\mathbf{c}(\mathbf{b}) \ll \mathbf{c}(\mathbf{A})$. However, as both $\mathbf{A}$ and $\mathbf{b}$ are part of the same linear constraints, their influence on the problem should be on a similar level.

In general, attribution methods struggle to capture the influence of \emph{combined parameters}, for example, the parameter $\mathbf{A}$, which encodes the graph structure via the incidence matrix, on MF and SP. Each element of $\mathbf{A}$ is not meaningful by itself, but only in combination with other elements. The gradient-based methods do not provide reasonable attributions for this parameter (Appendix \ref{sec:detailed_results}). These methods can only provide reasonable attributions for the edge weights.

Overall type I cases, IG only provides reasonable attributions with a near-zero baseline. The attributions generated by other baselines were either noisy or they were even misleading as shown in case MF1 in Tab. \ref{tab:res_t1}. At the baseline, all edges have the same capacity. However, the attributions indicate that only the last two edges are relevant, which is not reasonable, as all edges are saturated and thus equally important in this example. 

\begin{table}
\centering
\begin{subtable}{0.4\textwidth}
    \begin{tabular}{lcc}
    \toprule
    RO1 & $\mathbf{A}$ \& $\mathbf{c}(\mathbf{A})$ & $\mathbf{b}$ \& $\mathbf{c}(\mathbf{b})$\\
    \midrule
    Par & $\left(\begin{array}{cc}
    	1.0 & 1.0\\ 
    	2.0 & 1.0
        \end{array}\right)$ & $\left(\begin{array}{c}
        8.0 \\ 
        10.0
        \end{array}\right)$\\ 
    %    \midrule
    %RO1 & $\mathbf{c}(\mathbf{A})$ & $\mathbf{c}(\mathbf{b})$\\ \midrule
    Sal & $\left(\begin{array}{cc}
    	0.0 & -16.0\\ 
    	0.0 & 0.0
        \end{array}\right)$ & $\left(\begin{array}{c}
        2.0 \\
        0.0
        \end{array}\right)$\\ 
    GxI & $\left(\begin{array}{cc}
    	0.0 & -16.0\\ 
    	0.0 & 0.0
    \end{array}\right)$ & $\left(\begin{array}{c}
        16.0 \\
        0.0
        \end{array}\right)$\\ 
    \bottomrule
    \end{tabular}
    \caption{Case RO1}
\end{subtable}
\hfill
\begin{subtable}{0.55\textwidth}
    \begin{tabular}{lc}
    \toprule
    MF1 & $\mathbf{c}(\mathbf{b})$ \\
    \midrule
    Grad & $(0.0,\ 0.33,\ 0.33,\ 0.42,\ 0.09,\ 0.67,\ 0.59)^T$ \\
    GxI & $(0.0,\ 0.07,\ 0.20,\ 0.04,\ 0.04,\ 0.27,\ 0.29)^T$ \\
    IG-nz & $(0.0,\ 0.07,\ 0.20,\ 0.04,\ 0.04,\ 0.26,\ 0.27)^T$\\
    IG-as & $(0.0,\ 0.00,\ 0.00,\ 0.00,\ 0.00,\ 0.39,\ 0.49)^T$\\
    \bottomrule
    \end{tabular}
    \caption{Case MF1}
\end{subtable}

\caption{\textbf{(a)} The value of the parameters (Par) $\mathbf{b}$ are 8 times larger than the value of $\mathbf{A}$. This causes the saliency (Sal) attributions for $\mathbf{b}$ to be only 1/8 of the attributions for $\mathbf{A}$. The attributions for gradient times input (GxI) are more suitable in this case. \textbf{(b)} When comparing the attributions between a near-zero baseline for IG (IG-nz) and the baseline with the same values for all edges (IG-as), the latter are rather misleading, as they do not show any influence for all edges except the last two.}
\label{tab:res_t1}
\end{table}

\begin{figure*}
    \centering
    \begin{subfigure}[b]{0.3\textwidth}
        \centering
        \includegraphics[width=\textwidth]{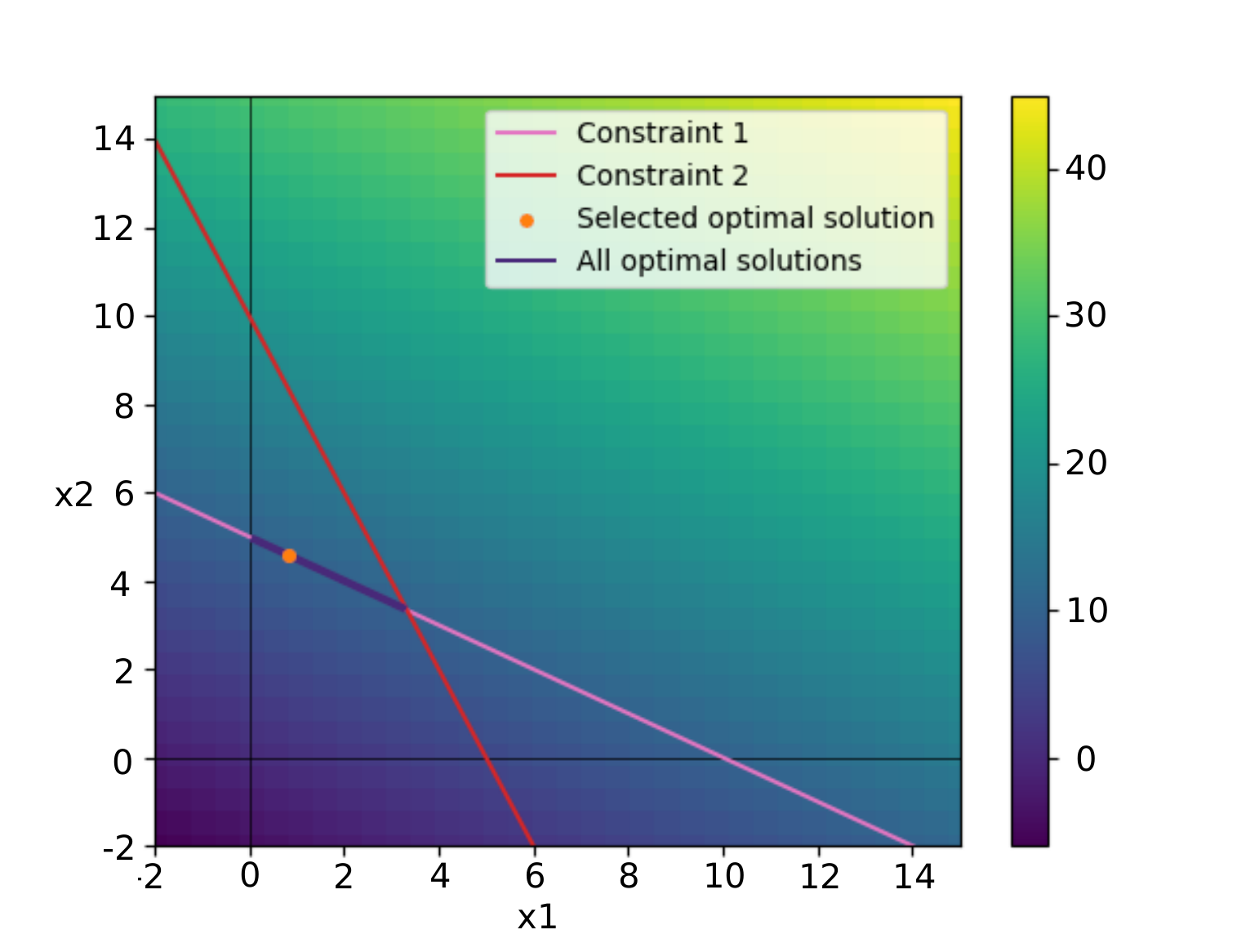}
        \label{fig:ro4_solutions}
        \caption{RO4 solutions}
    \end{subfigure}
    \begin{subfigure}[b]{0.32\textwidth}
        \centering
        \includegraphics[width=\textwidth]{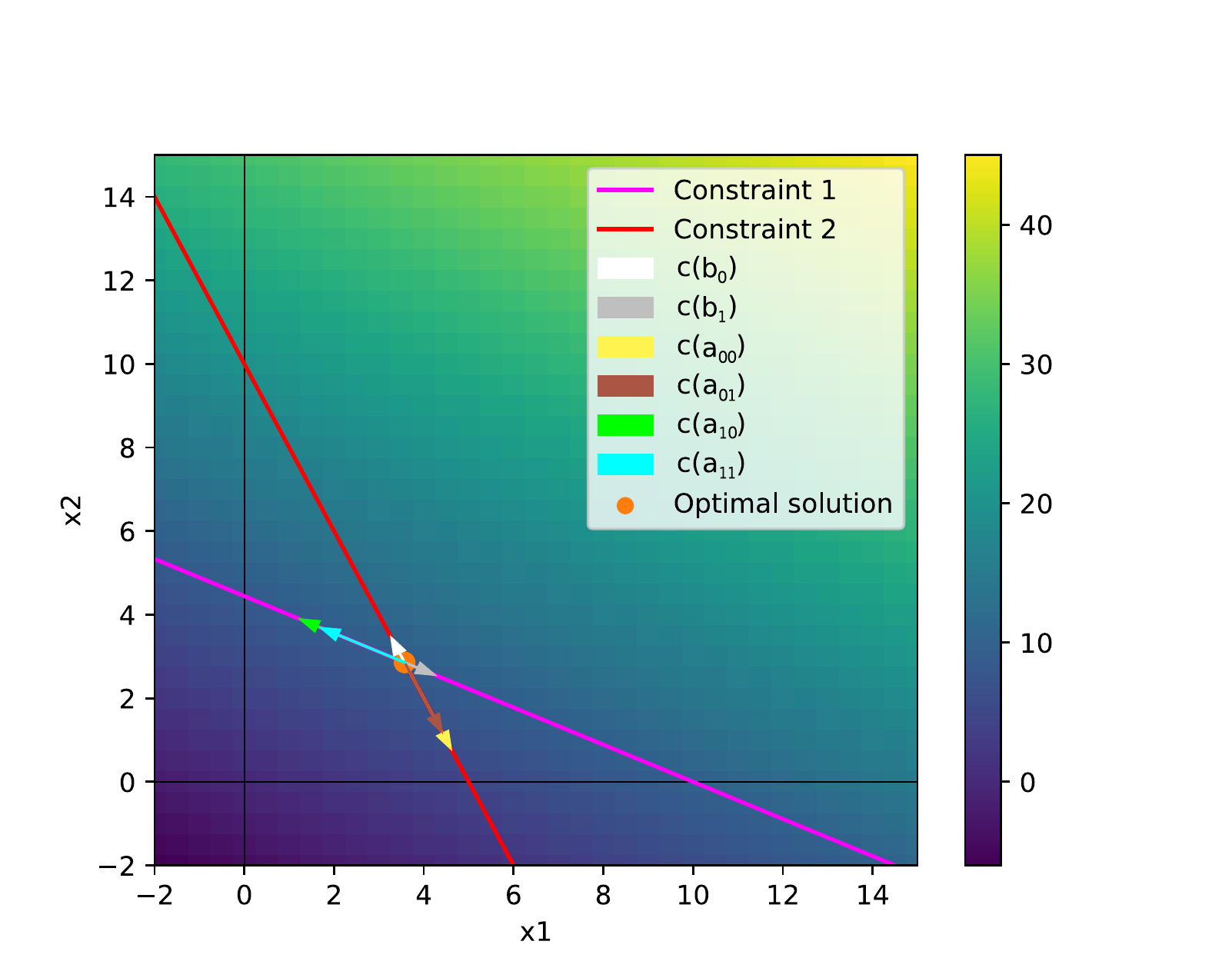}
        \label{fig:ro3_attributions}
        \caption{RO3 attributions}
    \end{subfigure}
    \begin{subfigure}[b]{0.32\textwidth}
        \centering
        \includegraphics[width=\textwidth]{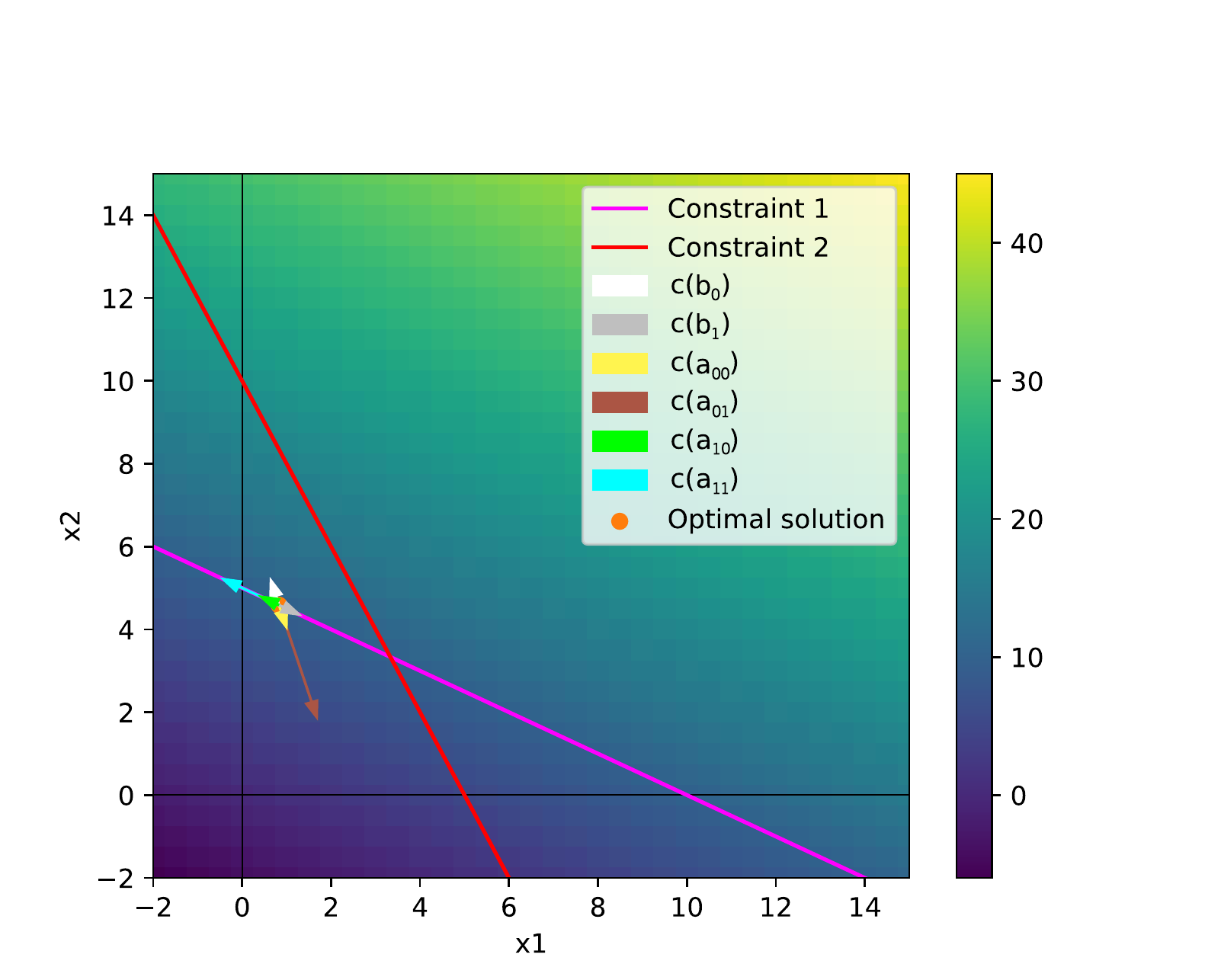}
        \label{fig:ro4_attributions}
        \caption{R04 attributions}
    \end{subfigure}
    \caption{\textbf{(a)} All possible optimal solutions for case RO4 with selected solution highlighted in orange. \textbf{(b)} and \textbf{(c)} Comparison of the attributions for the cases RO3 and RO4. At case RO3, there is a unique optimal solution, and the attributions provide reasonable information. For the case RO4, the attributions are still relatively similar to case RO3, even if here multiple optimal solutions are possible. The attributions are not able to provide information that multiple optimal solution exist.}
    \label{fig:problems_t2}
\end{figure*}

For type II cases, the attribution methods provided mixed results. In nearly all cases, the attributions for the solution map were \emph{not} able to highlight that there is more than one optimal solution. Instead, the attributions were similar to the corresponding type I cases. The reason for this behavior is that the LP solver just returns a single solution and does not provide any hints that more than one optimal solution might exist as shown in Fig. \ref{fig:problems_t2}. This behavior can be observed in all four problems. The objective map attributions have similar problems, except for RO4. Here, the attributions highlight that just one constraint is relevant, indicating that multiple optimal solutions exist. The attributions generated with Occ highlight that there is more than one optimal solution in all cases, however, they do not provide further insights which makes them not sufficient to explain such a situation. Otherwise, the observations for the type I cases about the saliency attribution scaling, attributions for combined parameters, and different baselines for IG are also true for the type II cases.

%\begin{figure*}
%    \centering
%    \begin{subfigure}[b]{0.33\textwidth}
%        \centering
%        \includegraphics[width=\textwidth]{problems_t2_left.pdf}
%        \label{fig:ro4_solutions}
%    \end{subfigure}
%    \hfill
%    \begin{subfigure}[b]{0.64\textwidth}
%        \centering
%        \includegraphics[width=\textwidth]{problems_t2_right.pdf}
%        \label{fig:ro4_attributions}
%    \end{subfigure}
%    \caption{\textbf{Left:} All possible optimal solution for case RO4. The solution of the solver is highlighted in orange. \textbf{Right:} Comparison of the attributions for the cases RO3 and RO4. At case RO3, there is a unique optimal solution, and the attributions provide reasonable information about the current situation. For the case RO4, the attributions are still relatively similar to case RO3, even if here multiple optimal solutions are possible. The attributions are not able to provide information that multiple optimal solution exist.}
%    \label{fig:problems_t2}
%\end{figure*}

The type III cases only consist of two different situations. In RO5, both constraints are active at the optimal solution and the solution is also limited by the implicit constraint $\mathbf{x} \geq 0$. As the implicit constraint is not defined by any parameters, the attribution methods fail, resulting in all methods providing misleading attributions for this case (see Appendix \ref{sec:detailed_results}).
The second type III example is KS3. Here, the normal set of items was extended by one very heavy item which cannot be selected, as its weight alone exceeds the knapsack limit. However, this item is very valuable. This causes misleading gradients for the solution map, and therefore also misleading attributions of the gradient-based methods (Tab. \ref{tab:res_t3}). Intuitively, it is expected that the attributions for the weight of all items have a -ve sign, as a larger weight, in general, corresponds to a reduced total value of the selected weights, because fewer items can be selected. However, the attributions for items 3 to 7 have a +ve sign, indicating the opposite connection which is not plausible. The attributions of Occ for this case were reasonable, as this method is not based on gradients.

% Overall, the type III cases are examples that specific situations exist in which the attribution methods do not provide reasonable attributions. This is most likely the case for any attribution method, however, one should keep in mind that attribution methods should not be trusted blindly. 

\begin{table}
\centering
\begin{tabular}{lc}
\toprule
KS3 & $\mathbf{c}(\mathbf{A})$ \\
\midrule
Sal & $(-0.08,\ -0.19,\ 0.04,\ 0.02,\ 0.04,\ 0.01,\ 0.03)^T$ \\
GxI & $(-1.65,\ -1.88,\ 0.18,\ 0.06,\ 0.24,\ 0.02,\ 0.10)^T$ \\
IG & $(-1.63,\ -1.86,\ 0.17,\ 0.06,\ 0.24,\ 0.02,\ 0.10)^T$ \\
\midrule
\midrule
& Attributions per item \\
Occ & $(0.00,\ 8.00,\ 0.00,\ 0.00,\ 0.00,\ 0.00,\ 0.00)^T$ \\
\bottomrule
\end{tabular}
\caption{This table shows the gradient-based attributions w.r.t.\ $\mathbf{A}$ for the objective map of case KS3. As $\mathbf{A}$ describes the item weights, there should be a -ve correlation between the values of A and the total value of the objective function. However, the attributions for some items have a +ve sign, indicating that increasing the weight of these items has a +ve influence on the total value, which is not reasonable. This behavior occurs for all three gradient-based methods. Occ is not affected by this (the Occ attributions show influence of a whole item on the result, which is always +ve or zero).}
\label{tab:res_t3}
\end{table}

{\bf Scaling Attributions - Real World Problem of Energy System Design.}
Real-world problems are often substantially more complex. This results in larger problems with more parameters, and therefore more attribution values. The increased number of values can make the attributions harder to understand. One example is an LP from the PlexPlain program \citep{PlexPlainTUD}, which models the energy level of a small house with a photovoltaic (PV) system. The full LP is shown in the supplement. As the energy level is modeled on an hourly basis for a year, it has over 40000 different problem variables. % Appendix \ref{sec:plexplain}. A way to make the attributions more manageable is to only consider a certain part of the problem, e.g., looking at the influence of one week.
By using Occ, the problem can be split into a small number of high-level meaningful structures, in this case, the months. In Tab. \ref{tab:plexplain}, the influence of each month on the battery capacity of the PV system is shown. It can be observed that the months in the spring and autumn have the largest influence. This is reasonable, as the battery is the most important during these seasons because it carries excess energy produced during the day into the evening. To explore the problem in greater depth, it would now be possible to select more fine-grained meaningful structures to look at.

\begin{table}
    \centering
    \begin{tabular}{lc@{\hspace{0.9\tabcolsep}}c@{\hspace{0.9\tabcolsep}}c@{\hspace{0.9\tabcolsep}}c@{\hspace{0.9\tabcolsep}}c@{\hspace{0.9\tabcolsep}}c@{\hspace{0.9\tabcolsep}}c@{\hspace{0.9\tabcolsep}}c@{\hspace{0.9\tabcolsep}}c@{\hspace{0.9\tabcolsep}}c@{\hspace{0.9\tabcolsep}}c@{\hspace{0.9\tabcolsep}}c@{\hspace{0.9\tabcolsep}}}
        \toprule
        Month & J & F & M & A & M & J & J & A & S & O & N & D \\
        Occ & 1 & 4 & \textcolor{green}{5} & \textcolor{green}{7} & 4 & \textcolor{orange}{0} & \textcolor{orange}{0} & \textcolor{orange}{0} & 3 & \textcolor{green}{4} & 1 & \textcolor{orange}{0} \\
        % Jan & 0.126\\
        % Feb & 0.486\\
        % Mar & 0.563\\
        % Apr & 0.709\\
        % May & 0.377\\
        % Jun & 0.032\\
        % Jul & 0.027\\
        % Aug & 0.042\\
        % Sep & 0.348\\
        % Oct & 0.428\\
        % Nov & 0.116\\
        % Dec &  0.048\\
        \bottomrule
    \end{tabular}
    \caption{Occ attributions show the influence of each month on the battery capacity. Large values (green) mean high influence, small values (orange) correspond to low influence.}
    \label{tab:plexplain}
\end{table}

\subsection{LP Attribution Methods Properties}\label{sec:properties}
For attribution methods, it is beneficial to have properties that describe the behavior of a method especially because it might be difficult to evaluate attributions for larger problems by hand. The following three properties are all proven to hold for Saliency, GxI, and IG (if they apply to that method). However, it turns out that they do not hold anymore for these methods when applied to LPs.

{\bf Sensitivity:} describes that the attributions are generated according to relations between input and output \citep{sundararahan2017axiomatic}. It can be split into two parts: If changing a model feature results in a change in the model output, that feature should have non-zero attributions. Additionally, if the output of a model remains the same despite a feature change, this feature should have zero attributions. This property is relevant for any kind of model and says that basic relations between input and output should be visible in the attributions. %It can be proved that IG has this property if used on NNs \citep{sundararahan2017axiomatic}.
As shown in Tab. \ref{tab:res_t1}, the first part of sensitivity is violated for the gradient-based methods. The attributions for the capacity of edge 1 are zero for all methods, but the edge has a large influence on the problem as it is responsible for $7/9$ of the total flow. The violation of the second part can be observed in Tab. \ref{tab:res_t3}. All methods have non-zero attributions for the first item, even if it does not affect the problem. Occ violates the first part of sensitivity in RO5. In theory, the second part should hold for Occ, as removing a non-relevant item from an LP should not change its solution. However, if such an item is removed and there were previously multiple optimal solutions, the solver may choose a different optimal solution, resulting in non-zero attributions. (See SP2, Appendix \ref{sec:detailed_results}).

{\bf Completeness:} also known as summation to delta \citep{shrikumar2017learning} only applies to attribution methods having a baseline. The attribution method fulfills completeness, if the condition $\sum{\mathbf{c}_\mathbf{M}(\mathbf{x}, \bar{\mathbf{x}})} = \mathbf{M}(\mathbf{x}) - \mathbf{M}(\bar{\mathbf{x}})$ is true i.e., attributions of a model at input $\mathbf{x}$ compared to a baseline $\bar{\mathbf{x}}$ should sum up to the difference between the model output at $\mathbf{x}$ and $\bar{\mathbf{x}}$. Again, IG fulfills this property for NNs, but not for LPs. 
% As this property is a strengthening of the first part of sensitivity. 

{\bf Implementation Invariance:} states that the attributions for two functionally equivalent, but differently implemented models should be the same \citep{sundararahan2017axiomatic}. This is reasonable, as the attributions should not depend on the specific implementation of the model, but rather on the function the model expresses. All considered attribution methods fulfill implementation invariance on NNs. Unfortunately, it can be shown that \emph{neither of these methods fulfills implementation invariance on LPs}. The reason behind this is that the solution of an LP, and therefore also its gradients and the attributions, sometimes depend on the LP solver. If an LP has multiple optimal solutions, the solver returns any of them. The returned solution depends not only on the type of solver but also on its implementation. Therefore, it can (and will) happen that two differently implemented solvers will return different optimal solutions resulting in different gradients leading to different attribution values for the same problem.
One could argue that the solver is technically not part of the model. Nevertheless, the attributions depend on the LP solver if the LP has more than one optimal solution and such cases can occur regularly \citep{koltai2000difference}. Therefore, this finding is important when attribution methods are applied to LPs. The solver dependence also prevents other properties of attributions methods on LPs. For e.g, Occ would fulfill the second part of sensitivity if the method would not be solver-dependent.

{\bf Further Properties:} There are other proposed properties which are not as relevant for LPs, as for example \emph{input invariance} \citep{kindermans2019reliability} or \emph{linearity}. These properties are especially desirable if the model has an NN-like structure and are therefore not as important for LPs.
%There are other properties for attribution methods suggested in the literature. Some of them are not relevant for LPs, as for example  This property states that the attributions should be invariant to a constant shift in the data, which is especially important for image-based neural networks, as these models can factor out such constant shifts. This property is not desirable for LPs, as a constant shift in the input data does change the LP and its solution. Therefore, the attributions should also change.
%\textit{Linearity} is another proposed property, which states that the attributions for two linearly composed models should be a linear composition of the attributions of the original models. This property is desirable for neural networks, as they consist partly of linear composition (or can be composed relatively easy). This property is not as important for LPs. Nevertheless, it is possible to prove that Occ fulfill linearity. For the gradient based attribution methods, a simple proof is not possible, as the attribution depend on the gradient of the LP, which is a nonlinear, unknown function. Therefore, proving or falsifying linearity for these methods is difficult. 
\begin{figure}[t]
    \begin{floatrow}
    \ffigbox{
      \includegraphics[width=0.3\textwidth]{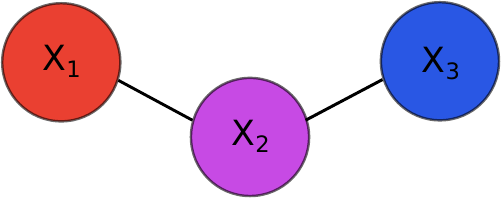}
    }{
      \caption{Example graph $G$. Here, $X_1$=1, $X_2$=0 and $X_3$ = equally likely 0 or 1.}
      \label{fig:map}
    }
    \capbtabbox{
    \hspace{1cm}
      \begin{tabular}{lccc}
        \toprule
        Occ & $X_1$ & $X_2$ & $X_3$ \\
        \midrule
        $E_{12}$ & 0 & -1 & -1 \\
        $E_{23}$ & 0 & 0 & -1 \\
        \bottomrule
      \end{tabular}
      \hspace{1cm}
    }{
        \caption{Occ attributions showing the influence of the edges on the variable states.}
      \label{tab:map}
    }
    \end{floatrow}
\end{figure}

\section{Implications of XLP}
If LPs were to be ``properly explainable" (refering to an intuitive, human level understanding), then this would have incredible implications for science and industry in general (well beyond artificial intelligence (AI) research). Climate-/energy systems researcher could design sustainable infrastructure to cover long-term energy demand \citep{schaber2012transmission}. However, LP explainability also naturally comes with strong implications within AI research directly due to established equivalences/use-cases like the discussed relationship between NNs and LPs or for instance in \emph{quantifying uncertainty and probabilistic reasoning} as in MAP inference for acquiring the ``most probable assignment",
\begin{align} \label{eq:map}
%\begin{split}
    &\arg \max_{\mathbf{x}\in\mathcal{X}} P(\mathbf{x};\pmb{\theta}) 
    =\arg \max_{\mathbf{x}\in\mathcal{X}} \frac{1}{Z(\pmb{\theta})}\exp(\langle \pmb{\theta},\phi(\mathbf{x})\rangle) 
    =\arg \max_{\pmb{\mu}\in\mathcal{M}} \langle \pmb{\theta}, \pmb{\mu}\rangle
%\end{split}
\end{align} 
Here $\mathcal{X}$ denotes the space of random variables $X_i$ and $P(\mathbf{x};\pmb{\theta})$ is the probability density function of a Markov Random Field (MRF) with parameters $\pmb{\theta}$, sufficient statistic $\phi$. Further $\mathcal{M}$ is called the \emph{marginal polytope} of the MRF's graph \citep{wainwright2008graphical} (formally, $\mathcal{M}(G)=\{\pmb{\mu}\mid\exists \pmb{\theta}\ldotp(\pmb{\mu}=\mathbb{E}[\phi(\mathbf{x})])\}$ with graph $G$, informally, the \emph{convex hull} of the vectors $\phi(\mathbf{x})$). It is known that several NP-hard combinatorial optimization problems can solved by exploiting the equivalence in Eq. \ref{eq:map} \citep{sontag2010approximate}, thereby, possibly allowing for a ``transportability" of explainability to such optimization problems (e.g.\ maximum cuts).

{\bf Attributions showcase:} Given graph $G$ in Fig. \ref{fig:map} which consists of three different random variables $X_1$, $X_2$ and $X_3$ each with binary states. $X_1$ is more likely to be 0, $X_3$ is more likely to be 1 and $X_2$ is equally likely in both states. The edges cause connected nodes to have more likely the same state. This results in all three variables having a zero state in the MAP. The exact values for $\phi$ and $\pmb{\theta}$ are presented in Appendix \ref{sec:map}. With the transformation to an LP shown in Eq. \ref{eq:map}, it is possible to compute attributions. Tab. \ref{tab:map} shows the Occ attributions for the edges in terms of the variable states instead of $\pmb{\mu}$, as $\pmb{\mu}$ only encodes the variable state, for a more condensed view. The attributions show that the edges have a negative influence on some nodes, causing them to be zero at MAP.

\section{Conclusions and Future Work}
We showed that generating attributions for linear programs with neural attribution methods is indeed possible, opening a new possible research area, XLP, to explain linear programs. Evaluation of four different attribution methods on several different LPs/ILPs showed that the methods have varying success. GxI and Occ provided overall the most reasonable attributions, especially in cases with a unique optimal solution. Saliency and IG were not as effective, because the former struggles with differently scaled input values and the baseline selection for the latter was ineffective. However, all methods struggled generally when LPs had more than one optimal solution. We showed that properties of attribution methods that are valid for NNs get \emph{violated} in LPs. 
As future work, solving the dependency of the attribution methods on the solver can make attribution methods for LPs apply better. The development of evaluation metrics for these methods is also an important task. Finally, designing better attribution methods specifically tailored to LPs is an important future direction. 

% To improve the IG method for LPs, it would be necessary to study possible baselines in more depth, possibly finding guidelines on how to select baselines for general LPs.

% Having a tool to compare different methods quantitatively can provide objective results, which is important.

%\section*{Acknowledgments}

    \bibliographystyle{plainnat}
\subsubsection*{Acknowledgments}
This work was supported by the ICT-48 Network of AI Research Excellence Center ``TAILOR'' (EU Horizon 2020, GA No 952215), the Nexplore Collaboration Lab ``AI in Construction'' (AICO) and by the Federal Ministry of Education and Research (BMBF; project ``PlexPlain'', FKZ 01IS19081). It benefited from the Hessian research priority programme LOEWE within the project WhiteBox and the HMWK cluster project ``The Third Wave of AI'' (3AI).
\bibliography{bibliography}

\clearpage
\appendix

\section{Appendix for ``Machines Explaining Linear Programs''}
We make use of this appendix following the main paper to provide additional details.

\section*{Code}
We have made our code publicly available at {\footnotesize \url{https://anonymous.4open.science/r/Machines-Explaining-Linear-Programs-75D2/README.md}}

\section{Problem specifications}
{\bf Maximum Flow Problem:} \label{sec:mf_descr}
For a given directed graph $G = (V, E)$ with nodes $V$ and edges $E$, the goal is to maximize the flow from a selected source node $s$ to the designated target node $t$. The problem can be described as follows:
\begin{equation}
 \setlength{\jot}{8pt}
	\begin{aligned}
		\text{maximize}  & \sum_{u: (u,t) \in E} f(u,t) && \\
		\text{subject to }  & f(u,v) \geq 0 & &\forall u, v \in V \\
		 - \sum_{u : (u,v) \in E}& f(u,v) + \sum_{w : (v,w) \in E} = 0 & &\forall v \in V \setminus \{s, t\} \\
		 &f(u,v) \leq c(u,v) & &\forall (u,v) \in E 
	\end{aligned}
 \end{equation}
 
In this formulation, $f(u,v)$ describes the flow from node $u$ to node $v$ and $c(u,v)$ is the capacity of the directed edge from $u$ to $v$. The first constraint ensures that the flow is always positive while the second constraint assures that the incoming and outgoing flow is the same for all nodes except source and target. The last constraint guarantees that the flow on each edge does not exceed the edge capacity. 

For the evaluation of this problem, the LP formulation is made more concise with the use of the \emph{incidence-matrix}. In the problem formulation, the optimization variable $\mathbf{x} \in \mathbb{R}^n$ describes the flow along each edge. The vector $\mathbf{w} \in \mathbb{R}^n$ is the negation of the target node row of the incidence matrix and describes which edges are connected to the target. The graph structure is encoded in $\mathbf{A} \in \mathbb{R}^{(m - 2)\times n}$ which holds all rows of the incidence matrix except the row for the source and the target node. The vector $\mathbf{b} \in \mathbb{R}^n$ describes the capacities for all edges. It is necessary that the ordering of edges in $\mathbf{A}$, $\mathbf{b}$, and $\mathbf{w}$ is the same. The full formulation is presented in Eq. \ref{eq:mf_std}.
\begin{equation} \label{eq:mf_std}
	\begin{aligned}
		&\text{maximize} & \mathbf{w}^T \mathbf{x} & \\
		&\text{subject to } & \mathbf{A} \mathbf{x} = 0 & \\
		& & \mathbf{x} \leq \mathbf{b} & \\ 
		& & \mathbf{x} \geq 0 &
	\end{aligned}
\end{equation}
        	
{\bf Shortest Path Problem:}\label{sec:sp_descr}
Given a directed graph $G = (V, E)$, with designated source node $s$ and target node $t$ as well as edge costs $w_{ij}$, the objective is to find the path which connects $s$ and $t$ with the lowest combined edge weights. The optimization variables $\mathbf{x}_{ij}$ for all edges $(i,j) \in E$ describe if the said edge is part of the optimal path or not. The optimization problem can then be formulated as follows:
\begin{equation}
	   \begin{aligned}
	    \text{minimize } & \sum_{(i, j) \in E} w_{ij} \mathbf{x}_{ij} &\\
		\text{subject to } & \forall i: \sum_{j} \mathbf{x}_{ij} - \sum_{j} \mathbf{x}_{ij} = & \begin{cases}
			1, & \text{ if } i = s;\\
			-1, & \text{ if } i = t;\\
			0, & \text{otherwise}
		\end{cases}\\
		&\mathbf{x}_{ij} \in \{0, 1\} &\forall (i, j) \in E
   \end{aligned}
\end{equation}

Similar to MF, it is possible to provide a more concise formulation for this problem as well. For this problem $\mathbf{A} \in \mathbb{R}^{m \times n}$ is the incidence-matrix. Next, $\mathbf{b} \in \mathbb{R}^n$ is a vector of all zeros, except for a 1 at the position of the source node and a -1 at the target node. Lastly, $\mathbf{w} \in \mathbb{R}^n$ is the vector of all edge costs. The ordering of the edges in $\mathbf{A}$, $\mathbf{b}$, and $\mathbf{w}$ has to be the same. This results in the following formulation:

\begin{equation} \label{eq:sp_std}
	\begin{aligned}
		\text{maximize } & & c^T \mathbf{x}  \\
		\text{subject to } & & A \mathbf{x} = b  \\
		\text{with } & & \mathbf{x} \geq 0  \\ 
		& & \mathbf{x} \in \{0, 1\}^n 
	\end{aligned}
\end{equation}

\begin{figure}
    \centering
    \includegraphics[scale=0.75]{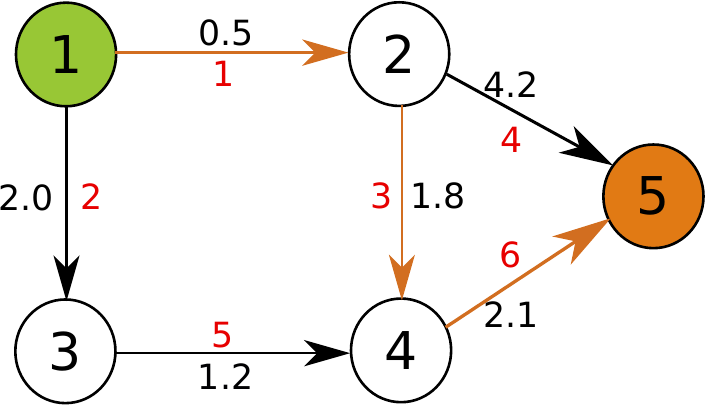}
    \caption{The graph for the shortest path problem. The source and target node are highlighted in green and orange respectively. The edge weights and the optimal solution are shown for SP1. The red numbers are unique edge identifiers.}
    \label{fig:sp_ex}
\end{figure}

The graph for SP is shown in Fig. \ref{fig:sp_ex}.

{\bf PlexPlain Problem:} \label{sec:plexplain}
The LP models the energy level of a simple house with a PV system and battery storage. Given the energy demand of the house over a year and the energy produced by a PV system, the capacity of the PV and the battery should be optimized. As the model spans a year and the energy levels are described on an hourly basis, the linear program consists of thousands of variables. The full LP formulation is given in Eq. \ref{eq:plexplain}
 
\begin{equation}\label{eq:plexplain}
\setlength{\jot}{8pt}
    \begin{aligned}
        \text{minimize} &&\text{co}_\text{PV} \text{ cap}_\text{PV} &+ \text{co}_\text{bat} \text{ cap}_\text{bat} + \text{co}_\text{buy} \sum_i \text{e}_{{buy}, i} &&\\
        \text{subject to}&& \text{e}_{\text{PV}, i} &\leq \text{cap}_\text{PV} \text{ available}_{\text{PV}, i}&&\\
             &&   \text{e}_{\text{bat}, i} &\leq \text{cap}_\text{bat}&&\\
             &&   \text{e}_\text{bat, start} &= \text{e}_\text{bat, end} - \text{e-out}_\text{start} + \text{e-in}_\text{start}&&\\
             &&   \text{e}_{\text{bat}, i} &= \text{e}_{\text{bat}, i - 1} - \text{e-out}_i + \text{e-in}_i \text{ } \forall i \geq 1 && \\
             &&   \text{demand}_i &= \text{e}_{{buy}, i} + \text{e-out}_i - \text{e-in}_i + \text{e}_{\text{PV}, i}&&
    \end{aligned}
\end{equation}
where ``co'' describes the cost and ``cap'' the capacity of an element. All variables with ``bat'' are related to the battery storage and ``buy'' is related to buying energy from the grid. All variables with ``e'' describe the energy of the element, for example $\text{e}_\text{PV}$ is the produced energy of the PV-system. The LP has $24 * 365 = 8760$ time steps and if not specified otherwise, $0 \leq i \le 8760$. 

{\bf Parameter values for all cases:} \label{sec:full_params}
In Tab. \ref{tab:param_ro}, \ref{tab:param_mf}, \ref{tab:param_ks} and \ref{tab:param_sp}, the full parameter values for all cases are provided. For MF the graph structure is provided in the main paper. The graph structure for SP is shown in Fig. \ref{fig:sp_ex}. The Parameters for these problems are given as information about the edges, as the general graph structure is the same for the different cases of these problems.
\begin{table}
    \centering
        \begin{tabular}{lccc}\toprule
        		 & $\mathbf{A}$ & $\mathbf{b}$ & $\mathbf{w}$\\ \midrule
        		RO1 & $\left(\begin{array}{cc}
        	1.0 & 1.0\\ 
        	2.0 & 1.0
        \end{array}\right)$ & $(8.0, 10.0)$ & $(1.0, 2.0)$\\ 
        		RO2 & $\left(\begin{array}{cc}
        	1.0 & 1.0\\ 
        	2.0 & 1.5
        \end{array}\right)$ & $(10.0, 10.0)$ & $(1.0, 2.0)$\\ 
        		RO3 & $\left(\begin{array}{cc}
        	1.0 & 2.25\\ 
        	2.0 & 1.0
        \end{array}\right)$ & $(10.0, 10.0)$ & $(1.0, 2.0)$\\ 
        		RO4 & $\left(\begin{array}{cc}
        	1.0 & 1.0\\ 
        	2.0 & 1.0
        \end{array}\right)$ & $(10.0, 10.0)$ & $(1.0, 2.0)$\\ 
        		RO5 & $\left(\begin{array}{cc}
        	1.0 & 2.0\\ 
        	2.0 & 1.0
        \end{array}\right)$ & $(10.0, 10.0)$ & $(1.0, 2.0)$\\
    	 \bottomrule
    	 \end{tabular}
	 \caption{Parameters for the resource-optimization problem.}
	 \label{tab:param_ro}
\end{table}

\begin{table}
	 \centering
	 \begin{tabular}{lSSSSSSS}\toprule
		Edge & 1 & 2 & 3 & 4 & 5 & 6 & 7\\ \midrule
		MF1 & 0.8 & 0.2 & 0.6 & 0.1 & 0.4 & 0.4 & 0.5\\ 
		MF2 & 0.8 & 0.2 & 0.6 & 0.3 & 0.4 & 0.2 & 0.3\\ 
	 \bottomrule
	 \end{tabular}
	 \caption{Edge capacities for the maximum flow problem cases. Both cases work with the same graph structure.}
	 \label{tab:param_mf}
\end{table}

\begin{table}
	 \centering
	 \begin{tabular}{lS@{\hspace{0.5\tabcolsep}}S@{\hspace{0.5\tabcolsep}}S@{\hspace{0.9\tabcolsep}}S@{\hspace{0.9\tabcolsep}}S@{\hspace{0.9\tabcolsep}}S@{\hspace{0.9\tabcolsep}}S|r}\toprule
		Item & 1 & 2 & 3 & 4 & 5 & 6 & 7 & $\mathbf{b}$\\ \midrule
		KS1 $\mathbf{w}$ & 3.0 & 2.0 & 4.0 & 2.0 & 2.0 &  &  & \\ 
		KS1 $\mathbf{A}$ & 5.0 & 3.0 & 6.0 & 2.0 & 4.0 &  &  & 10.0\\ 
		KS2 $\mathbf{w}$ & 4.0 & 4.0 & 1.0 & 3.0 &  &  &  & \\ 
		KS2 $\mathbf{A}$ & 4.0 & 6.0 & 3.0 & 3.0 &  &  &  & 10.0\\ 
		KS3 $\mathbf{w}$ & 50.0 & 15.0 & 3.0 & 2.0 & 4.0 & 2.0 & 3.0 & \\ 
		KS3 $\mathbf{A}$ & 20.0 & 10.0 & 5.0 & 3.0 & 6.0 & 2.0 & 4.0 & 10.0\\ 
	 \bottomrule
	 \end{tabular}
	 \caption{Parameters for the knapsack problem.}
	 \label{tab:param_ks}
\end{table}

\begin{table}
	 \centering
	 \begin{tabular}{lSSSSSS}\toprule
		Edge & 1 & 2 & 3 & 4 & 5 & 6\\ \midrule
		SP1 & 0.5 & 2.0 & 1.8 & 4.2 & 1.2 & 2.1\\ 
		SP2 & 0.5 & 2.0 & 1.8 & 3.9 & 1.2 & 2.1\\ 
	 \bottomrule
	 \end{tabular}
	 \caption{Edge weights for the shortest path problem cases. The graph structure is the same in both cases.}
	 \label{tab:param_sp}
\end{table}

{\bf Baselines for IG:} \label{sec:baselines_ig}
Several different baselines were tested for IG: As default value, the \emph{near-zero} baseline was used. The values of the baseline are all parameter values divided by 100. This is a relatively close approximation of a zero baseline, which is often not feasible in LPs. The other baselines were different between problems. For RO, three other baselines were tested, a baseline in which constraint 1 is the active constraint, one with constraint 2 active, and the last one with both constraints active. For MF and SP, a baseline with the same weights for all edges was used in addition to the near-zero baseline. For KS, a baseline with the average values and weights of all items (divided by 10) was used as well. The division by 10 is done to ensure that the baseline values are smaller than the values of the original input, otherwise, the attributions for the parameters would have different signs.

\section{Detailed Results} \label{sec:detailed_results}
{\bf Attributions for the incidence matrix:}
One of the situations where the gradient-based methods fail to produce reasonable attributions for parameters with combined meaning is the incidence matrix for the graph-based problems. The parameter $\mathbf{A}$ for the first case of the maximum flow problem is the following:.
\begin{equation}\label{eg:in_mat}
\mathbf{A} = \left( \begin{array}{lrrrrrrr}
        & e_1 & e_2 & e_3 & e_4 & e_5 & e_6 & e_7 \\
    n_2  & -1 & 0  & 1  & 1  & 0  & 0  & 0\\
    n_3  & 0  & -1 & -1 & 0  & 1  & 1  & 0\\
    n_4  & 0  & 0  & 0  & -1 & -1 & 0  & 1
    \end{array}\right)
\end{equation}
This parameter hold the full incidence-matrix, except the rows for the source and target nodes ($n_1$ and $n_5$, respectively). The saliency attributions w.r.t.\ $\mathbf{A}$ for the objective map are:
\begin{equation}
    \left(\arraycolsep=0.8pt\begin{array}{ccccccc}
	0.0 & 0.0 & 0.0 & 0.0 & 0.0 & 0.0 & 0.0\\ 
	-0.23 & -0.07 & -0.20 & -0.03 & -0.13 & -0.13 & -0.16\\ 
	-0.29 & -0.08 & -0.25 & -0.04 & -0.17 & -0.17 & -0.21
\end{array}\right)
\end{equation}
It does not seem reasonable that the attributions for the second node are all zero, as this node is responsible for $7/9$ of the total flow to the target node. Additionally, the attributions for the other two nodes are all negative, which is also not plausible. This or similar behavior occurs in all cases when attributions for the incidence matrix are computed.

\begin{figure}
    \centering
    \includegraphics[width=0.6\textwidth]{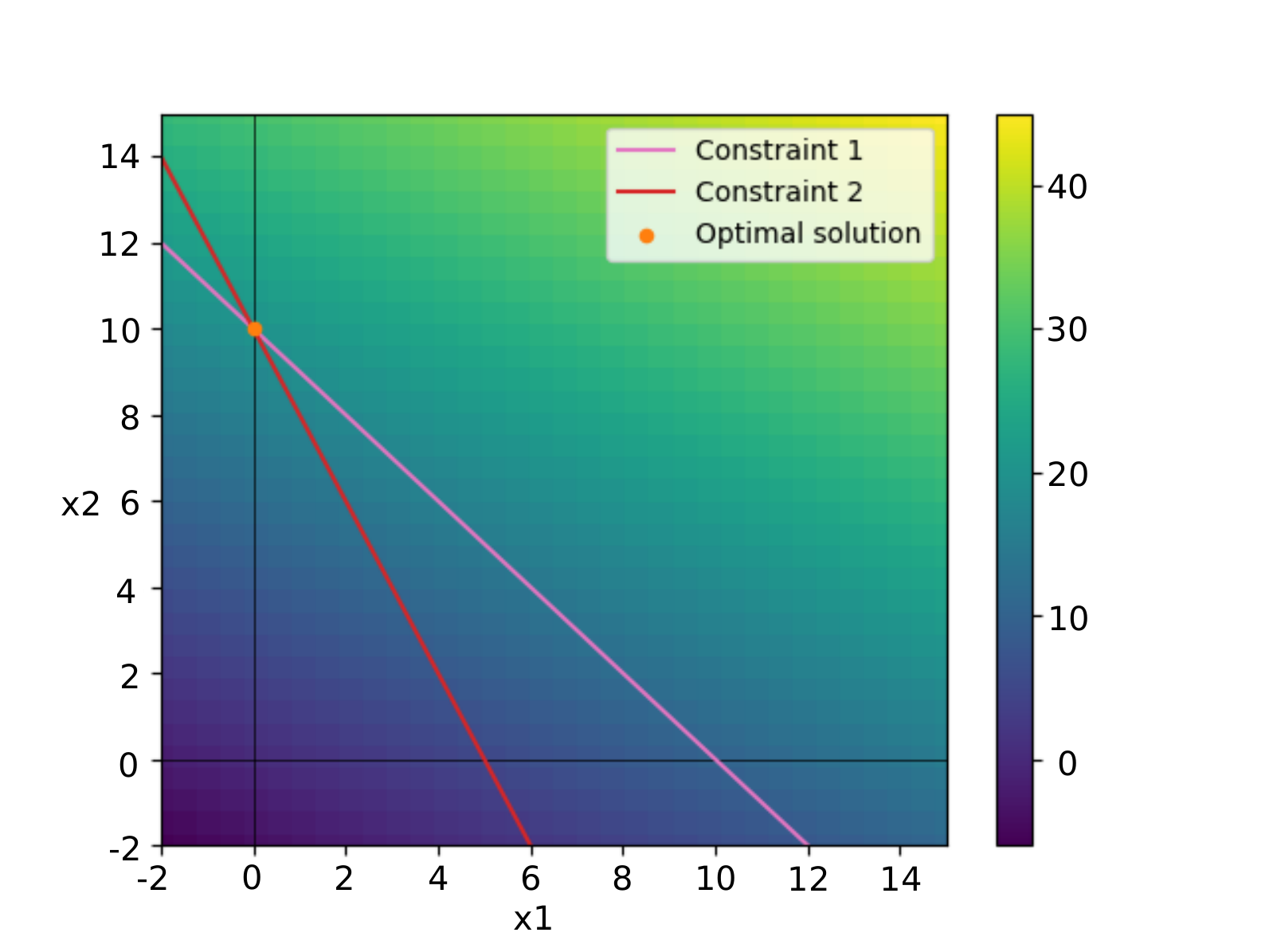}
    \caption{Visualization of RO5. The optimal solution is at the intersection of both constraints. Due to the limitation of $\mathbf{x} \geq 0$, the optimal solution is also at the $x_2$-axis.}
    \label{fig:RO5}
\end{figure}

{\bf More details about RO5:}
In this case, the problem has a unique optimal solution, which is at the intersection between both constraints and the $x_2$-axis (Fig. \ref{fig:RO5}). Both constraints and the implicit constraint $\mathbf{x} \geq 0$ are relevant. Additionally, the influence of both constraints should be similar, as they influence the optimal solution in the same way.

However, all gradient-based attribution methods show unbalanced influence for the parameters of both constraints. The results of Occ were even more misleading, as this method assigns zero attributions to both constraints. The results for the gradient-based methods are displayed in Tab. \ref{tab:RO5}.

{\renewcommand{\arraystretch}{1.2}
\begin{table}
    \centering
    \begin{tabular}{lcc}\toprule
        Method & $\mathbf{c}(\mathbf{A})$ & $\mathbf{c}(\mathbf{b})$ \\\midrule
        Grad & $\left(\begin{array}{cc}
            0.0 & -16.48 \\
            0.0 & -3.52
        \end{array}\right)$ & $\left(\begin{array}{c}
            1.65 \\
            0.35
        \end{array}\right)$ \\
        GxI & $\left(\begin{array}{cc}
            0.0 & -16.48 \\
            0.0 & -3.52
        \end{array}\right)$ & $\left(\begin{array}{c}
            16.48 \\
            3.52
        \end{array}\right)$ \\
        IG & $\left(\begin{array}{cc}
            0.0 & -12.50 \\
            0.0 & -7.30
        \end{array}\right)$ & $\left(\begin{array}{c}
            12.50 \\
            7.30
        \end{array}\right)$
        \\\bottomrule
    \end{tabular}
    \caption{Attributions for the objective map for RO5. It is expected that both constraints have a similar influence on the problem, which is not visible in the attributions. For IG, the results with the near-zero baseline are shown.}
    \label{tab:RO5}
\end{table}
}

{\bf Violations of Sensitivity:}
Occlusion does violate the sensitivity property on LPs, which can be observed in two different cases. The first part of sensitivity is violated in case RO5, where the method provides zero attributions for both constraints, even if both of them are relevant in this case (see Fig. \ref{fig:RO5}).

\begin{figure}
    \centering
    \includegraphics{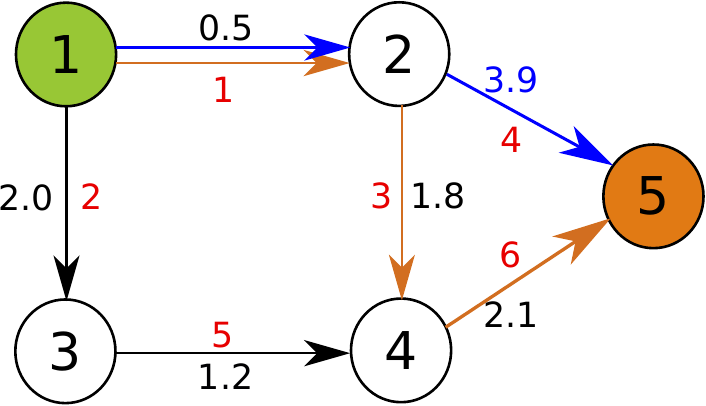}
    \caption{Graph for SP2. The source node is highlighted in green and the target node in orange. The first optimal path is shown in orange and the second one in blue. The edge numbers are shown in red.}
    \label{fig:SP2}
\end{figure}

The second part of sensitivity states, that the attributions for a parameter should be zero if that parameter does not influence the problem. One would assume that this property is valid for Occ, as removing a non-relevant part of the problem should not change the problem outcome, and therefore result in zero attributions. However, this is unfortunately not true. If a problem does have more than one optimal solution, the solver returns one of them. The attributions for the problem are based on this solution. If now a non-relevant part of the problem is removed, the solver may choose a different optimal solution, resulting in different attributions. This can be observed for SP2. The problem is shown in Fig. \ref{fig:SP2}, where both optimal paths are highlighted.
Neither edge 2 nor edge 5 does influence the optimal paths. However, the generated attributions for edges 2 and 5 are not zero, which violates the second part of sensitivity, as shown in Tab. \ref{tab:sp2_occ}.

\begin{table}
    \centering
    \begin{tabular}{lcccccc}
        \toprule
        Occ & $e_1$ & $e_2$ & $e_3$ & $e_4$ & $e_5$ & $e_6$ \\ \midrule
        $\mathbf{c}(e_2)$ & 0 & \text{none} & 1 & -1 & 0 & 1 \\
        $\mathbf{c}(e_5)$ & 0 & 0 & 1 & -1 & \text{none} & 1 \\
        \bottomrule
    \end{tabular}
    \caption{Occlusion attributions for case SP2.}
    \label{tab:sp2_occ}
\end{table}

\section{Applying Granger causality to LPs}\label{sec:gc_lp}
Suppose there is a set of parameters $\mathbf{X}$ and the model produces the output $\mathbf{Y}$. Then, a parameter $\mathbf{x}_i \in \mathbf{X}$ is causing $\mathbf{Y}$ by the definition of Granger, if $\epsilon_{\mathbf{X}} < \epsilon_{\mathbf{X} \setminus \{\mathbf{x}_i\}}$ \citep{granger1969investigating}. This means that the error of the model using the full parameter set $\mathbf{X}$ is smaller than the model using the full set except for $\mathbf{x}_i$. This has two assumptions: first, $\mathbf{X}$ contains all relevant information for $\mathbf{Y}$, and second, $\mathbf{X}$ is available before $\mathbf{Y}$. Both these assumptions are met for LPs.

As an LP does not have an inherent prediction error, the formulation cannot be applied directly. However, it is possible to use the model output as a ground truth. To get the difference in performance for an input parameter, it is removed from the LP. The modified program is solved and the difference between the output of the original and the modified model can be used as the Granger causal effect of the removed parameter. In the following equation, $\mathbf{X}$ is the combination of all parameters the linear program uses, and $\mathbf{x}_i$ is a certain parameter. The formulation can also be applied for the objective map, in this case, $\mathbf{S}$ has to be replaced by $\mathbf{O}$.

\begin{equation}\label{eq:gc}
   \mathbf{c}(\mathbf{x}_i) = \mathbf{S}(\mathbf{X}) - \mathbf{S}(\mathbf{X} \setminus \{\mathbf{x}_i\})
\end{equation}

This formulation is very close to the modified version of Occ for LPs.

\begin{figure}
    \centering
    \includegraphics{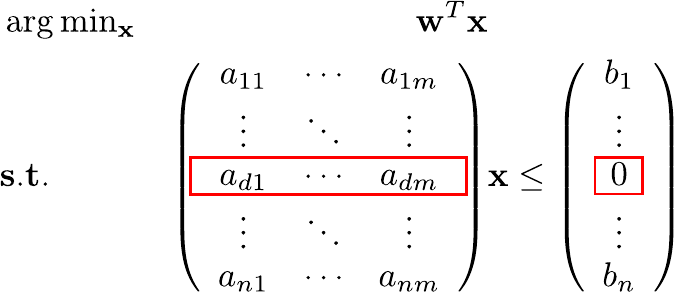}
    \caption{This figure shows the effects of masking a single element of one parameter from the problem. Setting a single entry of $\mathbf{b}$ to zero also affects the corresponding values of $\mathbf{A}$ (marked in red). If the full constraint is removed at once, no undesired side-effects occur and it is possible to compute attributions for the constraint. }
    \label{fig:lp_masking}
\end{figure}

\section{MAP assignment details}\label{sec:map}
The example for the MAP is based on a small graph which consists of the nodes $V = \{X_1, X_2, X_3\}$ and the edges $E = \{E_{12}, E_{23}\}$. Each node stands for one random variable which can have the states $\mathbf{x} \in \{0, 1\}$. The state of each random variable is influenced by the potentials $\phi_i$ for the nodes and $\phi_{ij}$ for the edges.
\begin{equation}
    \begin{aligned}
        \phi_{X_1}(0) &= 10 &\phi_{X_1}(0) &= 2 \\
        \phi_{X_2}(0) &= 8  &\phi_{X_2}(0) &= 8 \\
        \phi_{X_3}(0) &= 5  &\phi_{X_3}(0) &= 10 
    \end{aligned}
\end{equation}
\begin{equation}
    \phi_{E_{12}}(x_1, x_2) = \phi_{E_{23}}(x_2, x_3) = \begin{cases}
                20 & a = b \\
                1 & \text{otherwise}
            \end{cases}
\end{equation}
The parameters $\mathbf{\theta}_i = \log \phi_i$ and $\mathbf{\theta}_{ij} = \log \phi_{ij}$ are not normalized to obtain a proper probability distribution, because the normalization is not relevant for the LP version of the MAP. The LP formulation of the MAP problem works on the marginal polytope $\mathcal{M}$ of the Markov random field. This polytope is obtained by encoding the variable and edge states into a vector $\mu$ \citep{sontag2010approximate}. The resulting LP is then:

\begin{equation}
    \begin{aligned}
        \max_{\mu} \sum_{i \in V}\sum_{x_i}&\mathbf{\theta}_i(x_i)\mu_i(x_i) + \sum_{ij \in E}\sum_{x_i, x_j}\mathbf{\theta}_{ij}(x_i, x_j)\mu_{ij}(x_i, x_j)\\
        \vspace{0.5cm}\\
        \mu_i(x_i) &\in [0, 1] \text{ } \forall i \in V, x_i\\
        \sum_{x_i}\mu_i(x_i) &= 1 \text{ } \forall i \in V \\
        \mu_i(x_i) &= \sum_{x_j}\mu_{ij}(x_i, x_j) \text{ } \forall ij \in E, x_i\\
        \mu_j(x_j) &= \sum_{x_i}\mu_{ij}(x_i, x_j) \text{ } \forall ij \in E, x_j
    \end{aligned}
\end{equation}

In this LP, $\mu_i$ for $i \in V$ encodes the state of the random variable $i$. $\mu_{ij}$ encodes the state of edge $ij$. The constraints ensure that the encoding is correct. It is possible to use Occ to obtain attributions for parts of the problem. In this case, attributions were computed for both edges, by removing the respective parts of $\mu$ and $\mathbf{\theta}$ from the problem. The presented attributions in the main paper were obtained by reversing the encoding of the variable states from $\mu$ and performing the Occlusion formula on the variable states directly, which results in easier interpretable attributions.

\end{document}